# Reservoir Computing via Multi-Scale Random Fourier Features for Forecasting Fast–Slow Dynamical Systems

S. K. Laha

Advanced Design and Analysis Group

CSIR-Central Mechanical Engineering Research Institute

MG Avenue, Durgapur, West Bengal, PIN-713209, India

**Abstract:**

Forecasting nonlinear time series with multi-scale temporal structures remains a central challenge in complex systems modeling. We present a novel reservoir computing framework that combines delay embedding with random Fourier feature (RFF) mappings to capture such dynamics. Two formulations are investigated: a single-scale RFF reservoir, which employs a fixed kernel bandwidth, and a multi-scale RFF reservoir, which integrates multiple bandwidths to represent both fast and slow temporal dependencies. The framework is applied to a diverse set of canonical systems: neuronal models such as the Rulkov map, Izhikevich model, Hindmarsh–Rose model, and Morris–Lecar model, which exhibit spiking, bursting, and chaotic behaviors arising from fast–slow interactions; and ecological models including the predator–prey dynamics and Ricker map with seasonal forcing, which display multi-scale oscillations and intermittency. Across all cases, the multi-scale RFF reservoir consistently outperforms its single-scale counterpart, achieving lower normalized root mean square error (NRMSE) and more robust long-horizon predictions. These results highlight the effectiveness of explicitly incorporating multi-scale feature mappings into reservoir computing architectures for modeling complex dynamical systems with intrinsic fast–slow interactions.

## I. Introduction:

Multi-scale temporal dynamics are a defining feature of complex systems, where interactions between fast and slow processes generate rich behaviors such as oscillations, bursting, intermittency, and chaos. Kuehn [1] provides a comprehensive foundation for analyzing such systems, framing multiple timescale dynamics as an extension of singular perturbation theory and surveying asymptotic techniques like averaging and matched expansions. In biological contexts, Murray [2] shows how physiological and ecological processes are governed by coupled scales, from rapid biochemical reactions to long-term population dynamics, while Lei [4] highlights circadian rhythms and gene regulation as canonical examples of slow–fast interactions. Broader reviews by Vlachos [3], Walpole et al. [5], and Meier-Schellersheim et al. [6] emphasize that multiscale modeling spans biology, physics, materials, and engineering, requiring hybrid simulation methods, coarse-graining, and data-driven analysis to bridge microscopic processes with macroscopic behaviors. Weinan [7] formalizes these links through homogenization, renormalization, and the Mori–Zwanzig framework, providing a unified mathematical treatment, while Scheffer and Carpenter [8] underscore their ecological importance by connecting slow environmental forcing to sudden regime shifts in ecosystems. Together, these works establish that multi-scale interactions are not domain-specific but rather universal, and that capturing both rapid fluctuations and slow modulatory trends is crucial for prediction and control.

In neuroscience, these principles are concretely embodied in canonical models. The Rulkov map [9] abstracts spiking–bursting behavior in a discrete fast–slow system. The Izhikevich model [10] achieves biologically plausible spiking and bursting using a fast voltage variable and a slower recovery term. The Hindmarsh–Rose model [11] introduces an additional slow adaptation current to generate complex bursting. The Morris–Lecar model [12] links voltage oscillations to calcium and potassium dynamics with distinct timescales. In ecology, the Ricker map [13] captures fast density-dependent fluctuations modulated by slow environmental forcing. These canonical systems illustrate the ubiquity of timescale separation, providing challenging benchmarks for predictive models.

Traditional mechanistic approaches to multi-scale dynamics rely on singular perturbation theory [1] or hybrid simulations [3], but often face limitations when governing equations are unknown or high-dimensional. This has motivated the rise of machine-learning approaches, particularly reservoir computing (RC), which projects inputs into a high-dimensional dynamical reservoir and trains only a linear readout [14–16]. RC avoids the challenges of backpropagation through time while retaining the ability to model nonlinear and chaotic time series. Its success spans chaotic attractors, turbulence, and climate dynamics [20–24]. Pathak et al. [20, 21] showed RC's predictive power for large spatiotemporal chaotic systems and hybrid schemes with mechanistic models. Lu et al. [22] reconstructed attractors with RC while preserving long-term ergodic behavior, and Chattopadhyay et al. [23] demonstrated its superiority over deep learning methods for multi-scale Lorenz-96 dynamics.

Recent advances have extended RC to explicitly address multi-scale prediction. Tanaka et al. [31] introduced heterogeneous leaky integrator neurons to capture dynamics at multiple timescales. Zheng et al. [30] proposed long–short term ESNs with multi-reservoir structures, while Inoue et al. [41] optimized leaking rates in deep ESNs for multi-scale dynamics. Lin et al. [35, 37] developed multi-scaling RC for noise-induced transitions, while Tsuchiyama et al. [36] analyzed the role of temporal resolution in chaotic reproduction. These works highlight the importance of multiscale structure in reservoir design.

Despite these advances, a systematic framework for explicitly encoding fast–slow dynamics in kernelized reservoir computing remains missing. While heterogeneous leaking rates and deep reservoirs approximate multiple timescales, they do not directly enforce multiscale structure in the feature space. To address this gap, we propose two novel RC variations based on Random Fourier Features (RFF): a Single-Scale formulation using a global bandwidth, and a Multi-Scale formulation with variable-specific bandwidths. The theoretical foundation of Random Fourier Features (RFF) was established by Rahimi and Recht [42], who showed that shift-invariant kernels can be efficiently approximated by mapping data into a randomized low-dimensional feature space. This approach transforms nonlinear kernel methods into linear learning problems. By tailoring the kernel mapping to fast and slow subsystems, the Multi-Scale approach explicitly embeds temporal separation into the reservoir. We benchmark both models on six canonical fast–slow systems — Rulkov [9], Izhikevich [10], Hindmarsh–Rose [11], Morris–Lecar [12], Ricker [13], and Predator–Prey — evaluating one-step-ahead and closed-loop multi-step predictions. Our results demonstrate that while Single-Scale RFF-RC provides competitive short-term accuracy, Multi-Scale RFF-RC consistently improves long-horizon stability, establishing multiscale feature design as a critical advancement in reservoir computing for nonlinear time-series forecasting.

## II. Methods

### A. Problem Formulation

Consider a $d$-dimensional dynamical system generating time series observations $\{\mathbf{u}(t)\}_{t=1}^{T}$ where $\mathbf{u}(t) = [u_1(t), u_2(t), \ldots, u_d(t)]^T \in \mathbb{R}^d$. The system evolves according to an unknown discrete-time map:

$$\mathbf{u}(t+1) = \mathbf{F}(\mathbf{u}(t), \mathbf{u}(t-1), \ldots, \mathbf{u}(t-k+1)) \tag{1}$$

where $\mathbf{F}: \mathbb{R}^{dk} \to \mathbb{R}^d$ represents the unknown system dynamics and $k$ is the memory length.

Our objective is to learn a predictor $\hat{\mathbf{F}}: \mathbb{R}^{dk} \to \mathbb{R}^d$ that approximates the true dynamics using delay-embedded observations. This enables both one-step-ahead and multi-step autonomous forecasting.

#### 1. Delay Embedding Construction

To create supervised training pairs, we employ lagged vectors of past samples. For a chosen lag order $k \geq 1$, define for $t = k+1, \ldots, T$

$$\mathbf{v}_t^{(i)} = [u_i(t-k), u_i(t-k+1), \ldots, u_i(t-1)]^T \in \mathbb{R}^k \tag{2}$$

which collects the most recent $k$ values of component $i$.

Concatenating across variables yields the composite delay vector is given by

$$\mathbf{U}_t = [(\mathbf{v}_t^{(1)})^T, (\mathbf{v}_t^{(2)})^T, \ldots, (\mathbf{v}_t^{(d)})^T]^T \in \mathbb{R}^{dk} \tag{3}$$

The corresponding one-step target is simply

$$\mathbf{Y}_t = \mathbf{u}(t) \in \mathbb{R}^d \tag{4}$$

Stacking training pairs for $t = k+1, \ldots, T$ gives the feature matrix $\mathbf{Z}$ and the target matrix $\mathbf{Y}$:

$$\mathbf{Z} = \begin{bmatrix} \Phi(\mathbf{U}_{k+1})^T \\ \Phi(\mathbf{U}_{k+2})^T \\ \vdots \\ \Phi(\mathbf{U}_T)^T \end{bmatrix}, \mathbf{Y} = \begin{bmatrix} \mathbf{u}(k+1)^T \\ \mathbf{u}(k+2)^T \\ \vdots \\ \mathbf{u}(T)^T \end{bmatrix} \quad (5)$$

where $\Phi(\cdot)$ is the chosen RFF mapping and $n = T - k$.

## 2. Random Fourier Features Framework

Random Fourier Features (RFF) [42] provide computationally efficient approximations to shift-invariant kernels via Bochner's theorem. For the Gaussian RBF kernel:

$$\kappa(\mathbf{x}, \mathbf{y}) = \exp\left(-\frac{\|\mathbf{x} - \mathbf{y}\|^2}{2\sigma^2}\right)$$

Bochner's theorem guarantees the spectral representation:

$$\kappa(\boldsymbol{\delta}) = \int_{\mathbb{R}^{dk}} e^{i\boldsymbol{\omega}^T \boldsymbol{\delta}} p(\boldsymbol{\omega}) d\boldsymbol{\omega}$$

where $p(\boldsymbol{\omega}) = (2\pi\sigma^2)^{-dk/2}\exp(-\sigma^2\|\boldsymbol{\omega}\|^2/2)$ is the Fourier transform of the Gaussian kernel.

### a. Single-Scale RFF construction

The single-scale approach employs uniform bandwidth $\sigma$ across all variables and dimensions.

- Sample frequency matrix: $\mathbf{W} \in \mathbb{R}^{dk \times m}$ with entries $w_{ij} \sim \mathcal{N}(0, \sigma^{-2})$
- Sample phase vector: $\mathbf{b} \in \mathbb{R}^m$ with entries $b_j \sim \mathcal{U}[0, 2\pi]$

The feature mapping is then

$$\boldsymbol{\phi}_{\text{single}}(\mathbf{U}_t) = \sqrt{\frac{2}{m}} \cos(\mathbf{W}^T \mathbf{U}_t + \mathbf{b}) \in \mathbb{R}^m \quad (6)$$

This yields feature matrix $\mathbf{Z}_{\text{single}} \in \mathbb{R}^{n \times m}$ with $M = m$.

### b. Multi-Scale RFF Construction

The multi-scale approach recognizes that different system variables may exhibit distinct characteristic time scales and dynamical properties, requiring variable-specific kernel bandwidths.

For each component $i \in \{1, 2, \dots, d\}$:

- Select bandwidth $\sigma_i > 0$ and feature count $m_i > 0$

- Sample frequency matrix: $\mathbf{W}^{(i)} \in \mathbb{R}^{k \times m_i}$ with entries $\mathcal{N}(0, \sigma_i^{-2})$
- Sample phase vector: $\mathbf{b}^{(i)} \in \mathbb{R}^{m_i}$ with entries $\mathcal{U}[0, 2\pi]$

The block feature map is then constructed as follows

$$\boldsymbol{\phi}^{(i)}(\mathbf{v}_t^{(i)}) = \sqrt{\frac{2}{m_i}} \cos((\mathbf{W}^{(i)})^T \mathbf{v}_t^{(i)} + \mathbf{b}^{(i)}) \in \mathbb{R}^{m_i} \quad (7)$$

Concatenating blocks across variables yields the multi-scale feature representation

$$\boldsymbol{\phi}_{\text{multi}}(\mathbf{U}_t) = [\boldsymbol{\phi}^{(1)}(\mathbf{v}_t^{(1)})^T, \ldots, \boldsymbol{\phi}^{(d)}(\mathbf{v}_t^{(d)})^T]^T \in \mathbb{R}^M \quad (8)$$

where $M = \sum_{i=1}^{d} m_i$ is the total feature dimension.

This construction allows each variable to be processed with its optimal bandwidth while maintaining the block structure that reflects the system's natural variable decomposition.

### 3. Ridge Regression

Given feature matrix $\mathbf{Z} \in \mathbb{R}^{n \times M}$ and targets $\mathbf{Y} \in \mathbb{R}^{n \times d}$, multi-output ridge regression with regularization parameter $\lambda > 0$ solves

$$\min_{\mathbf{W}} \|\mathbf{Y} - \mathbf{Z}\mathbf{W}\|_F^2 + \lambda \|\mathbf{W}\|_F^2 \quad (9)$$

where $\mathbf{W} \in \mathbb{R}^{M \times d}$. The closed-form solution is

$$\mathbf{W}^* = (\mathbf{Z}^T \mathbf{Z} + \lambda \mathbf{I}_M)^{-1} \mathbf{Z}^T \mathbf{Y} \quad (10)$$

The regularization parameter $\lambda$ controls the bias-variance trade-off: small $\lambda$ causes low bias, high variance with potential overfitting; whereas large $\lambda$ may cause high bias, low variance and underfitting. The optimal $\lambda$ is typically selected via cross-validation.

### 4. Forecasting

*Single-step prediction:* For a new delay vector $\mathbf{U}_{\text{test}}$,

$$\hat{\mathbf{u}}(t) = \boldsymbol{\phi}(\mathbf{U}_{\text{test}})^T \mathbf{W}^* \quad (11)$$

where $\boldsymbol{\phi} \in \{\boldsymbol{\phi}_{\text{single}}, \boldsymbol{\phi}_{\text{multi}}\}$ depending on the chosen approach.

*Multi-step closed-loop forecasting:* Starting with an initial delay $\mathbf{U}_{t_0}$, iterate for $h = 1, \ldots, H$:

1. Predict $\hat{\mathbf{u}}(t_0 + h) = \boldsymbol{\phi}(\mathbf{U}_{t_0+h-1})^T \mathbf{W}^*$

2. Update the delay by removing the oldest lag and appending $\hat{\mathbf{u}}(t_0 + h)$, preserving the structure of (1)–(2)

3. Repeat until horizon $H$.

**III. Results:**

To systematically evaluate the effectiveness of Multi-Scale Random Fourier Features (RFF-RC) compared to the conventional Single-Scale approach, we applied both frameworks to a set of canonical fast–slow dynamical models drawn from neuroscience and ecology. These models, summarized in Table 1, represent well-known testbeds that capture fundamental mechanisms of timescale separation: fast subsystems describing rapid membrane potential oscillations or prey population dynamics, and slow subsystems governing recovery, adaptation, or environmental forcing. The selected systems span discrete-time maps (Rulkov, Ricker), reduced biophysical models (Morris–Lecar, Hindmarsh–Rose), an efficient spiking formulation (Izhikevich), and an ecological predator–prey system. This diversity allows us to test whether the Multi-Scale RFF-RC framework generalizes across domains while retaining its capacity to resolve sharp transients in fast variables and gradual modulations in slow variables. In the following subsections, we present quantitative comparisons of one-step-ahead and closed-loop multi-step predictions, highlighting the conditions under which Multi-Scale RFF-RC offers substantial gains in accuracy and robustness over Single-Scale RFF-RC.

To compare predictive accuracy across models, we use the Normalized Root Mean Square Error (NRMSE), a scale-independent metric that quantifies the deviation between the predicted trajectory $\hat{y}_t$ and the true trajectory $y_t$. For a time series of length $T$, the NRMSE is defined as:

$$\text{NRMSE} = \frac{\sqrt{\frac{1}{T}\sum_{t=1}^{T}(y_t - \hat{y}_t)^2}}{max(y_t) - min(y_t)} \qquad (12)$$

Table: Fast-Slow Dynamic Models

| Model | Type | Variables | Fast–Slow Assignment | Biophysical / Ecological Interpretation | Dynamics Captured |
|---|---|---|---|---|---|
| Rulkov Map | Discrete-time, 2D map | $x, y$ | Fast: $x$ (membrane potential); Slow: $y$ (recovery) | Abstract neuronal map capturing excitability with time-scale separation | Spiking, bursting, chaos |
| Morris–Lecar | Biophysical, 2D ODE | $V, n$ | Fast: $V$ (membrane voltage); Slow: $n$ (K$^+$ gating) | Reduced conductance-based model with calcium and potassium dynamics | Spiking, bistability, oscillations |
| Hindmarsh–Rose | Abstract biophysical, 3D ODE | $x, y, z$ | Fast: $x$ (membrane potential), $y$ (fast recovery); Slow: $z$ (adaptation) | Phenomenological model of neurons with fast spikes and slow bursting | Spiking, bursting, chaos |
| Izhikevich | Abstracted 2D ODE + reset | $v, u$ | Fast: $v$ (membrane potential); Slow: $u$ (recovery) | Hybrid quadratic integrate-and-fire model with adaptation mechanism | Wide range of spiking & bursting regimes |
| Ricker Map | Discrete-time, 2D map | $x, r$ | Fast: $x$ (population density); Slow: $r$ (growth rate / forcing) | Ecological map with seasonal forcing and density-dependent regulation | Population cycles, chaos |
| Predator–Prey | Continuous-time, 2D ODE | $x, y$ | Fast: $x$ (prey); Slow: $y$ (predator) | Ecological interaction model with predator–prey oscillations and delay | Stable cycles, oscillations |

### A. Rulkov Map

The predictive performance of the proposed Single-Scale and Multi-Scale Random Fourier Features (RFF) Reservoir Computing (RC) framework was evaluated on the Rulkov neuronal map [9]. This map is a two-dimensional discrete-time system that captures realistic spiking–bursting behavior with low computational cost. It is defined by

$$x_{n+1} = \frac{\alpha}{1-x_n} + y_n$$
$$y_{n+1} = y_n - \mu(x_n + 1) + \mu\sigma \quad (12)$$

(12) where $\alpha = 4.1, \mu = 0.001, \sigma = -1.6$. The fast variable $x_n$ represents the membrane potential and is responsible for rapid voltage oscillations. The slow variable $y_n$ governs recovery processes such as ionic gating and adaptation. Because $\mu \ll 1$, the system exhibits alternating fast spikes and slow bursts, and depending on parameters it can also show chaotic behavior. This makes prediction difficult, as the model must capture both sharp nonlinear excursions and gradual long-term drifts.

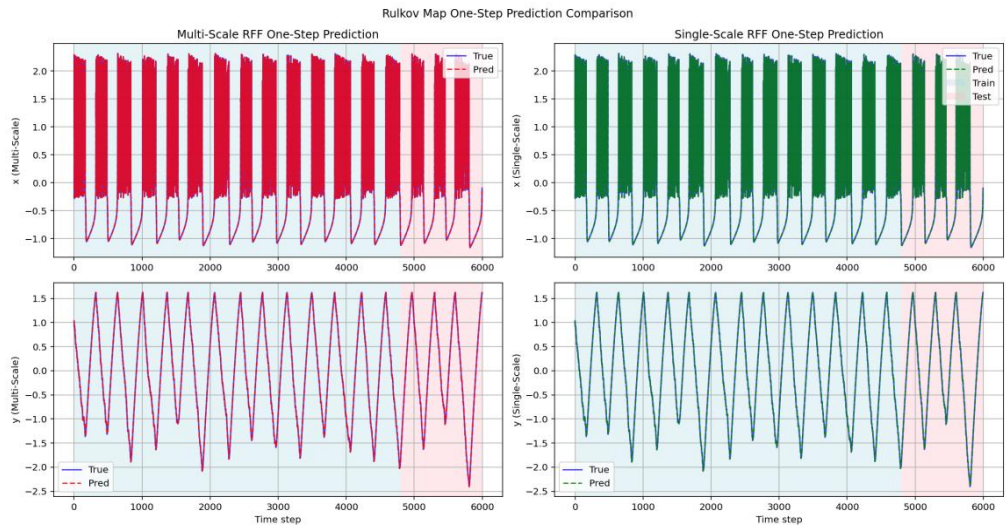

**FIG.1:** One-Step Prediction of Rulkov Map Using Reservoir Computing with Single- and Multi-Scale Random Fourier Features

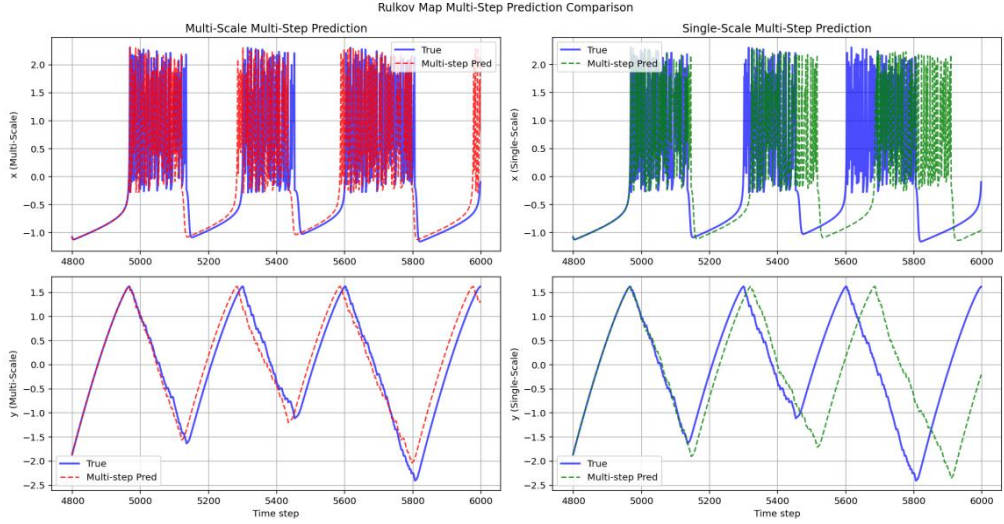

**FIG. 2:** Multi-Step Closed Loop Prediction of Rulkov Map Using Reservoir Computing with Single- and Multi-Scale Random Fourier Features

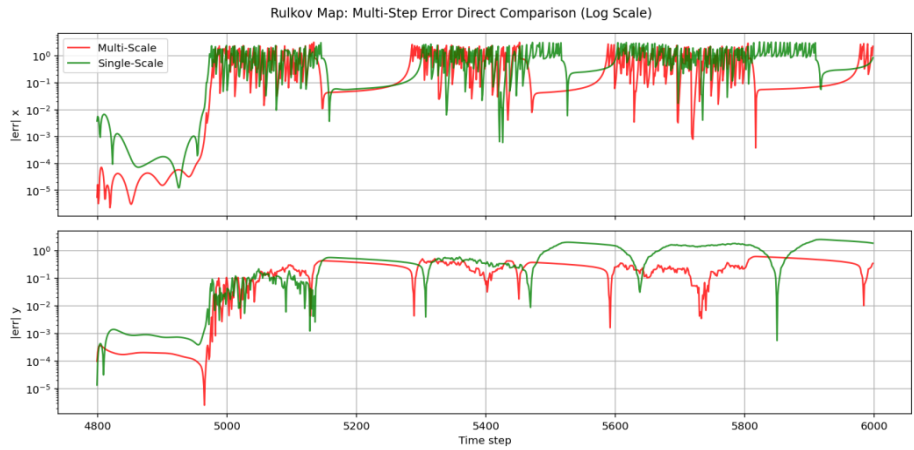

**FIG. 3:** Error Growth in Long-Horizon Predictions: Log-Scale Comparison of Single- and Multi-Scale Random Fourier Feature Reservoirs for Rulkov Map

In one-step-ahead prediction experiments, results using Multi-Scale RFF-RC and Single-Scale RFF-RC are shown in Fig. 1. In this experiment we use variable-specific bandwidths for RFF which are $\sigma_x = 0.1, \sigma_y = 10.0$. The Multi-Scale RFF-RC achieved very low normalized root mean square errors: $\text{NRMSE}_x = 2.61 \times 10^{-5}$ and $\text{NRMSE}_y = 1.92 \times 10^{-5}$. The Single-Scale RFF-RC, using a single global bandwidth, had much higher errors: $\text{NRMSE}_x = 1.60 \times 10^{-3}$ and $\text{NRMSE}_y = 1.70 \times 10^{-4}$. Thus, Multi-Scale RFF-RC improved accuracy by nearly two orders of magnitude for the fast subsystem and by an order of magnitude for the slow subsystem.

For multi-step closed-loop prediction (Fig. 2), the performance gap remained clear. Errors accumulated over long horizons due to the system's inherent chaotic nature, but Multi-Scale consistently provided lower error levels. Both Single-Scale and Multi-Scale RFF-RC result in phase lag with respect to the actual ground truth. But in Multi-Scale RFF-RC it is less prominent.

Direct error trajectories (Fig. 3) further highlight these differences. Multi-Scale reduced short-term fluctuations and delayed divergence, particularly during bursting episodes. Single-Scale errors grew more rapidly and displayed larger variance across cycles. These results demonstrate that a single global kernel bandwidth cannot capture the dual time scales of the Rulkov map. By contrast, Multi-Scale RFF-RC uses variable-specific bandwidths to create a richer feature space. This allows it to preserve fast transients in the membrane potential while also modeling slow recovery dynamics. As a result, the Multi-Scale framework provides more accurate and robust predictions for neuronal fast–slow systems.

**B. Hindmarsh–Rose model**

The Hindmarsh–Rose (HR) model [11] was next used to assess the predictive ability of Multi-Scale RFF-RC and Single-Scale RFF-RC. This model is a three-dimensional dynamical system that captures a wide variety of neuronal firing patterns, particularly spiking and bursting activity driven by fast–slow interactions. It is described by the system of equations

$$\frac{dx}{dt} = y - ax^3 + bx^2 - z + I$$
$$\frac{dy}{dt} = c - dx^2 - y \qquad (13)$$
$$\frac{dz}{dt} = r\left(s(x - x_R) - z\right)$$

where $x$ is the membrane potential, $y$ is a recovery variable associated with fast ionic currents, and $z$ is a slow adaptation current that regulates bursting dynamics. The parameters $a, b, c, d$ control the fast subsystem, while $r, s, x_R$ govern the slow adaptation process. Biologically, this separation of timescales allows the HR model to reproduce tonic spiking, periodic bursting, and chaotic bursting, depending on the external current $I$ and the slow adaptation rate $r$. For example, with $a = 1.0, b = 3.0, c = 1.0, d = 5.0, r = 0.01, s = 4.0, x_R = -1.6$, the system generates bursting with alternating epochs of fast spikes and quiescent phases. Increasing $r$ shortens bursts, while tuning $I$ transitions the system between quiescence, tonic firing, and bursting regimes.

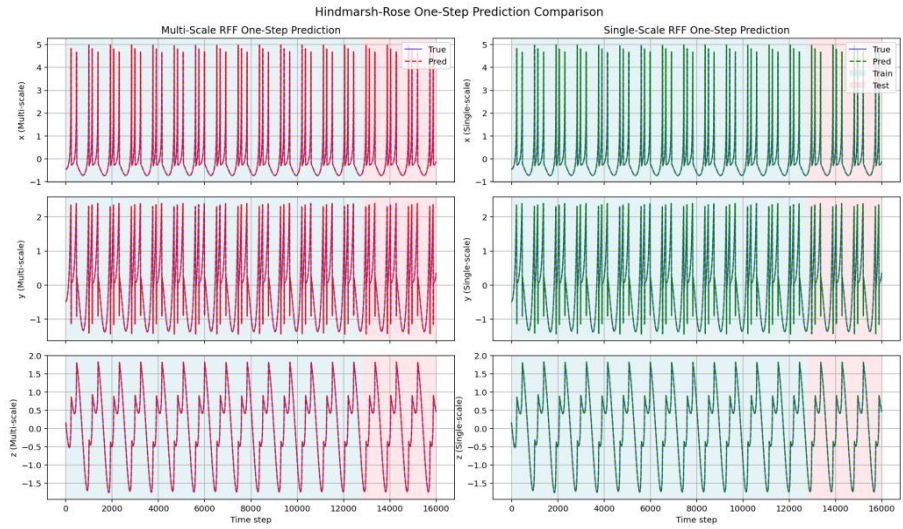

**FIG.4:** One-Step Prediction of Hindmarsh–Rose model Using Reservoir Computing with Single- and Multi-Scale Random Fourier Features

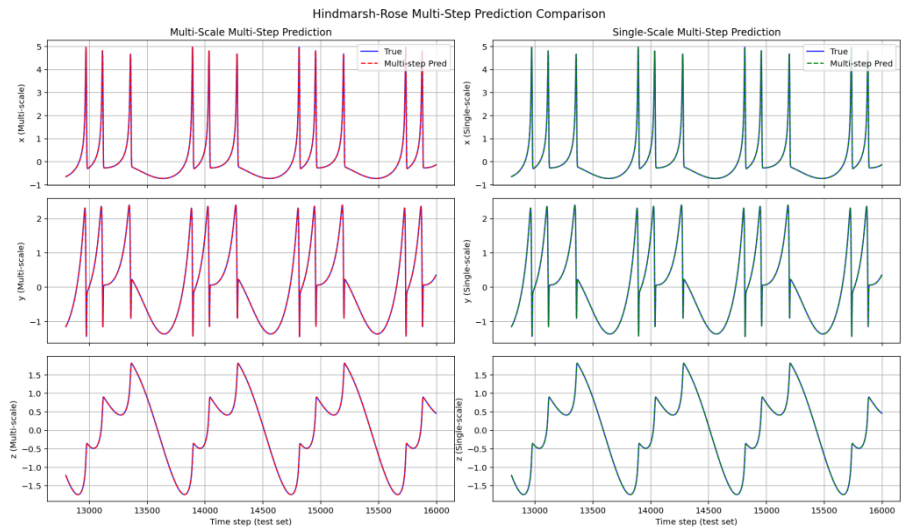

**FIG. 5:** Multi-Step Closed Loop Prediction of Hindmarsh–Rose model Using Reservoir Computing with Single- and Multi-Scale Random Fourier Features

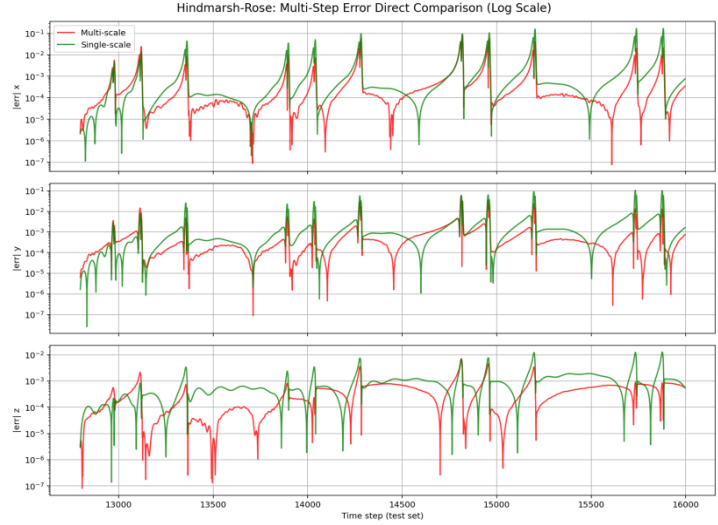

**FIG. 6:** Error Growth in Long-Horizon Predictions: Log-Scale Comparison of Single- and Multi-Scale Random Fourier Feature Reservoirs for Hindmarsh–Rose model

For one-step-ahead prediction (Fig. 4), both Multi-Scale and Single-Scale RFF-RC achieved extremely low errors, reflecting the relatively smooth trajectory of the system. However, Multi-Scale RFF-RC produced consistently better results: $\text{NRMSE}_x = 1.29 \times 10^{-6}$, $\text{NRMSE}_y = 2.19 \times 10^{-6}$, $\text{NRMSE}_z = 2.38 \times 10^{-6}$, compared to Single-Scale values of $\text{NRMSE}_x = 2.85 \times 10^{-6}$, $\text{NRMSE}_y = 2.89 \times 10^{-6}$, $\text{NRMSE}_z = 3.07 \times 10^{-6}$. This improvement was achieved using variable-specific bandwidths $[0.1, 1.0, 10.0]$ for $(x, y, z)$, which enabled better representation of both the fast subsystem $(x, y)$ and the slow adaptation variable $z$.

The difference became clearer in closed-loop multi-step prediction (Fig. 5). Here, Multi-Scale RFF-RC maintained lower error growth across all variables, with $\text{NRMSE}_x = 1.84 \times 10^{-3}$, $\text{NRMSE}_y = 1.76 \times 10^{-3}$, $\text{NRMSE}_z = 3.60 \times 10^{-4}$. In comparison, Single-Scale RFF-RC produced larger errors: $\text{NRMSE}_x = 2.41 \times 10^{-3}$, $\text{NRMSE}_y = 2.31 \times 10^{-3}$, $\text{NRMSE}_z = 4.69 \times 10^{-4}$. While both methods remained stable over long horizons, Multi-Scale consistently slowed error accumulation and more accurately reproduced the bursting envelope and adaptation dynamics.

The log-scale absolute error trajectories in Fig. 6 quantify this disparity across the test horizon. For all variables, multi-scale errors (green) remain confined below $10^{-4}$ for the majority of the sequence, occasionally spiking to $10^{-3}$ during burst onsets but decaying rapidly thereafter. Single-scale errors (red), however, escalate more frequently to $10^{-2} - 10^{-1}$ levels, especially in $y$ and $z$ variables, reflecting amplified sensitivity to initial perturbations. Overall, the HR results show that Multi-Scale RFF-RC offers superior predictive accuracy and robustness, particularly in capturing the multi-timescale interplay between fast spiking and slow adaptation that defines bursting behavior.

**C. Izhikevich model**

The Izhikevich model [10] was also examined as a benchmark for evaluating the Single-Scale and Multi-Scale RFF-RC frameworks. This model provides a computationally efficient yet biologically meaningful description of neuronal excitability and can reproduce a wide range of firing patterns, from regular spiking to bursting, depending on parameter choice. It is defined by the two-dimensional system

$$\frac{dv}{dt} = 0.04v^2 + 5v + 140 - u + I$$
$$\frac{du}{dt} = a(bv - u) \tag{14}$$

with an auxiliary reset condition: if $v \geq 30$, then $v \leftarrow c$ and $u \leftarrow u + d$. Here, $v$ represents the membrane potential, while $u$ is a recovery variable that accounts for ionic currents and adaptation. The applied current $I$ drives excitation, and the parameters $a, b, c, d$ control the dynamics. Specifically, $a$ determines the time scale of recovery, $b$ sets the sensitivity of recovery to subthreshold voltage, $c$ is the reset value of the membrane potential after a spike, and $d$ represents the after-spike increment of the recovery variable. Despite its simplicity, the Izhikevich model can generate tonic spiking, phasic spiking, bursting, and chaotic firing patterns. In this study, we considered the tonic bursting regime with parameter values $a = 0.02, b = 0.2, c = -50, d = 2$ under constant input current $I = 10$.

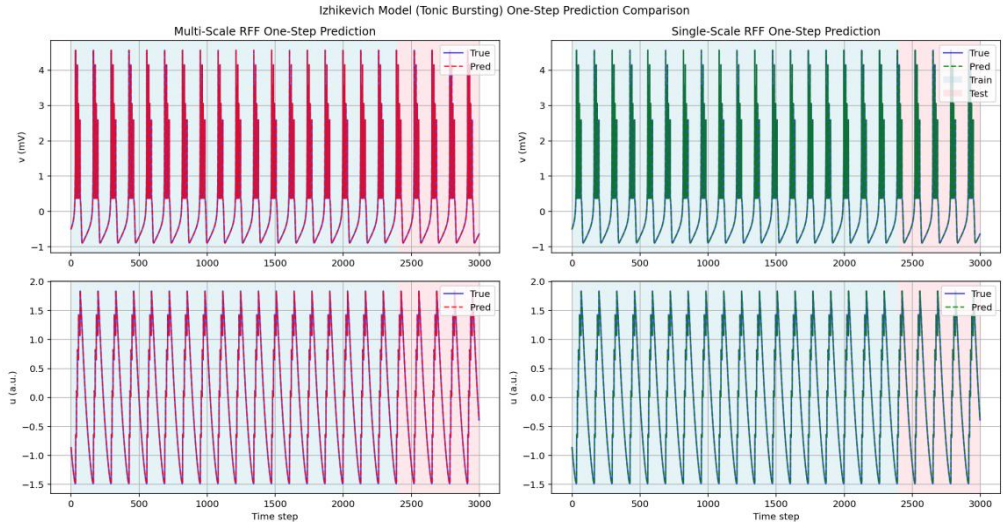

**FIG.7:** One-Step Prediction of Izhikevich model Using Reservoir Computing with Single- and Multi-Scale Random Fourier Features

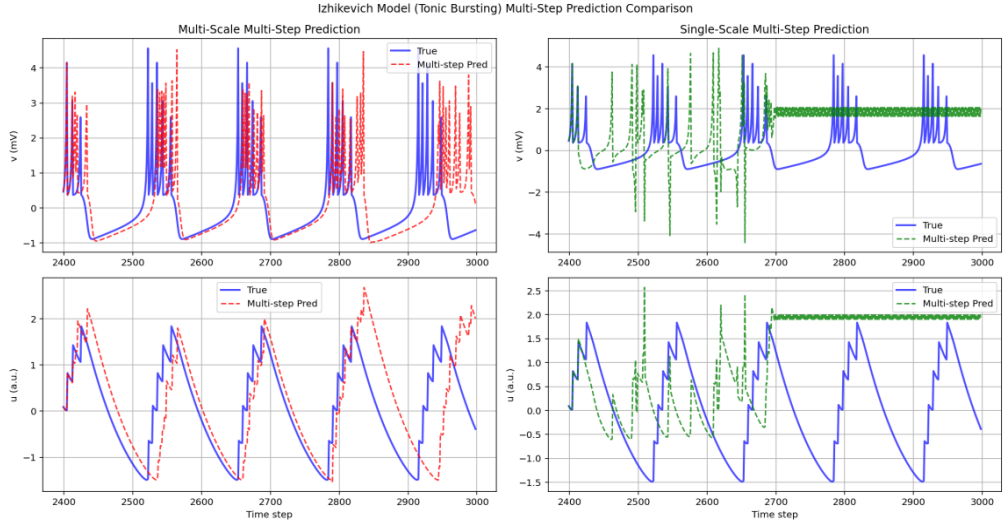

**FIG. 8:** Multi-Step Closed Loop Prediction of Izhikevich model Using Reservoir Computing with Single- and Multi-Scale Random Fourier Features

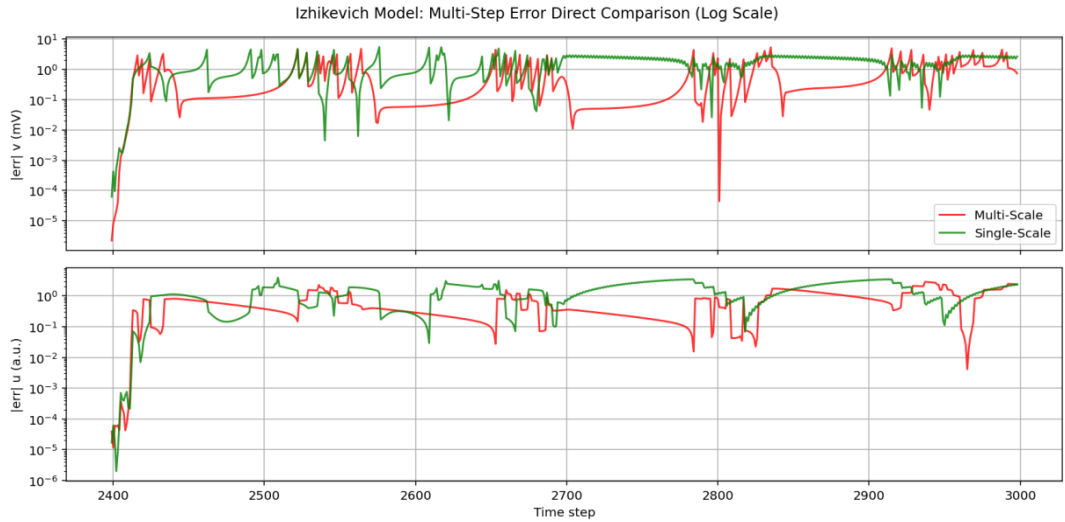

**FIG. 9:** Error Growth in Long-Horizon Predictions: Log-Scale Comparison of Single- and Multi-Scale Random Fourier Feature Reservoirs for Izhikevich model

Figure 7 shows that both reservoir variants perform well in one-step ahead forecasting. In the multi-scale RFF-RC (left panel, red), predicted trajectories almost perfectly overlap the ground truth (blue), faithfully reproducing spike amplitudes and inter-burst intervals. With bandwidths tuned separately for the fast voltage ($\sigma_v = 0.1$) and the slower recovery variable ($\sigma_u = 10.0$), Multi-Scale produced $\text{NRMSE}_v = 1.84 \times 10^{-6}$ and $\text{NRMSE}_u = 1.23 \times 10^{-5}$. The Single-Scale RFF-RC, by contrast, had larger errors of $\text{NRMSE}_v = 6.90 \times 10^{-5}$ and $\text{NRMSE}_u = 2.10 \times 10^{-5}$.

In multi-step closed-loop prediction (Fig. 8), the advantage of Multi-Scale RFF-RC became more evident. Multi-scale predictions sustain the correct bursting rhythm with small phase drift or amplitude decay, whereas single-scale predictions diverge very quickly.

Fig. 9 compares multi-step prediction errors for the Izhikevich model, with the fast membrane potential $v$ shown above and the slow recovery variable $u$ below. The single-scale RFF-RC (green) exhibits larger and more persistent errors, particularly during spiking episodes. In contrast, multi-scale RFF-RC (red) maintains lower and more transient errors, thereby preserving spike timing and recovery trends over the full test interval. These results confirm the superior stability of the multi-scale approach for long-horizon forecasting of fast–slow neuronal dynamics. Overall, the multi-Scale RFF-RC consistently captured both components, achieving superior accuracy in one-step prediction and far greater robustness in closed-loop forecasting of tonic bursting activity.

**D. Predator–Prey model**

The predator–prey system was also analyzed as a representative ecological fast–slow model to evaluate the Multi-Scale Random Fourier Features (RFF) framework. The classical formulation is a modified Lotka–Volterra system with timescale separation, given by

$$\begin{aligned} \frac{dx}{dt} &= x(\alpha - \beta y) \\ \frac{dy}{dt} &= \epsilon y(\delta x - \gamma) \end{aligned} \tag{15}$$

where $x$ denotes the prey population and $y$ the predator population. The parameters $\alpha$ and $\beta$ represent prey growth and predation rates, while $\delta$ and $\gamma$ denote predator efficiency and mortality. The parameter $\epsilon \ll 1$ imposes a slower timescale on predator dynamics relative to prey, creating a fast–slow interaction. In this setting, prey evolve quickly, driven by intrinsic reproduction and predation, while predator populations adjust gradually according to prey availability. Such systems generate oscillations resembling ecological cycles, with prey rising first, followed by delayed predator growth, and eventual decline of both. This delayed coupling and separation of timescales make the system an important benchmark for testing multi-scale predictive models.

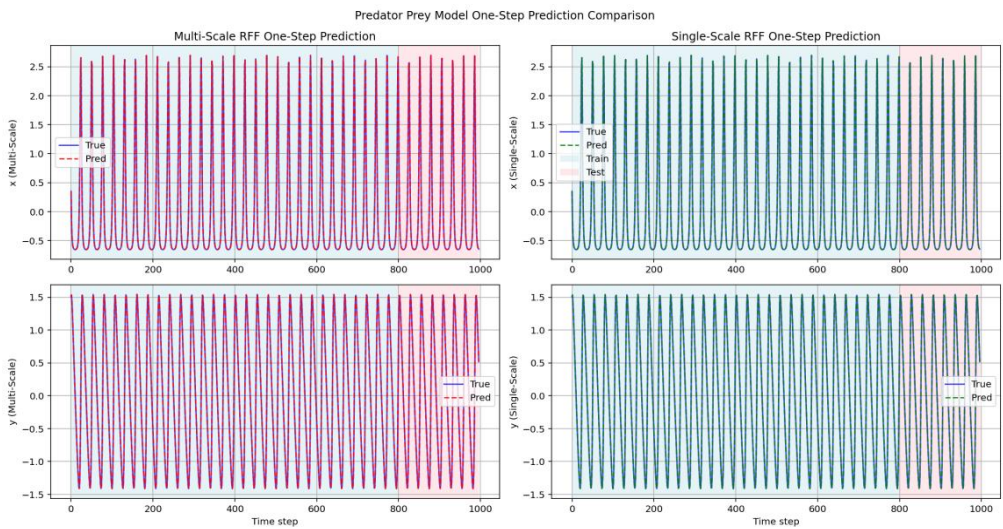

**FIG. 10:** One-Step Prediction of predator–prey model Using Reservoir Computing with Single- and Multi-Scale Random Fourier Features

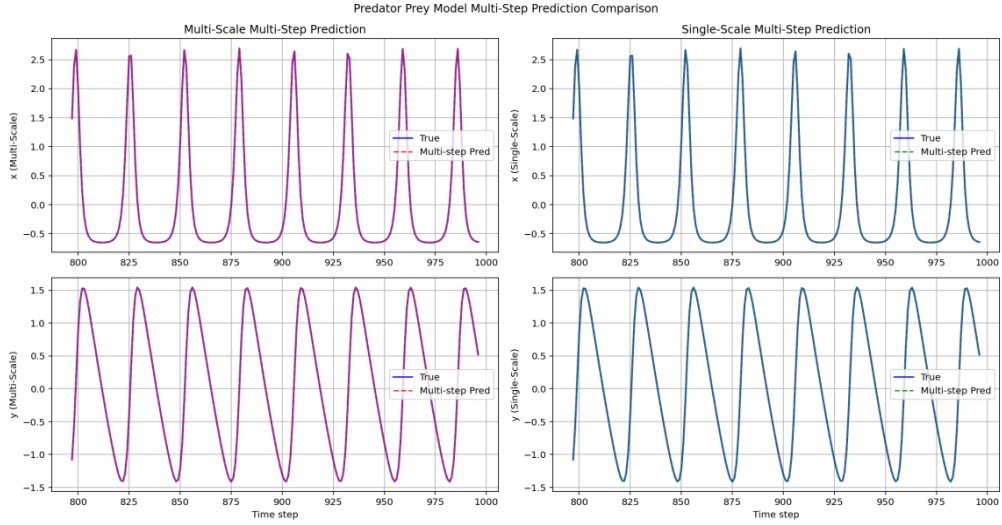

**FIG. 11:** Multi-Step Closed Loop Prediction of predator–prey model Using Reservoir Computing with Single- and Multi-Scale Random Fourier Features

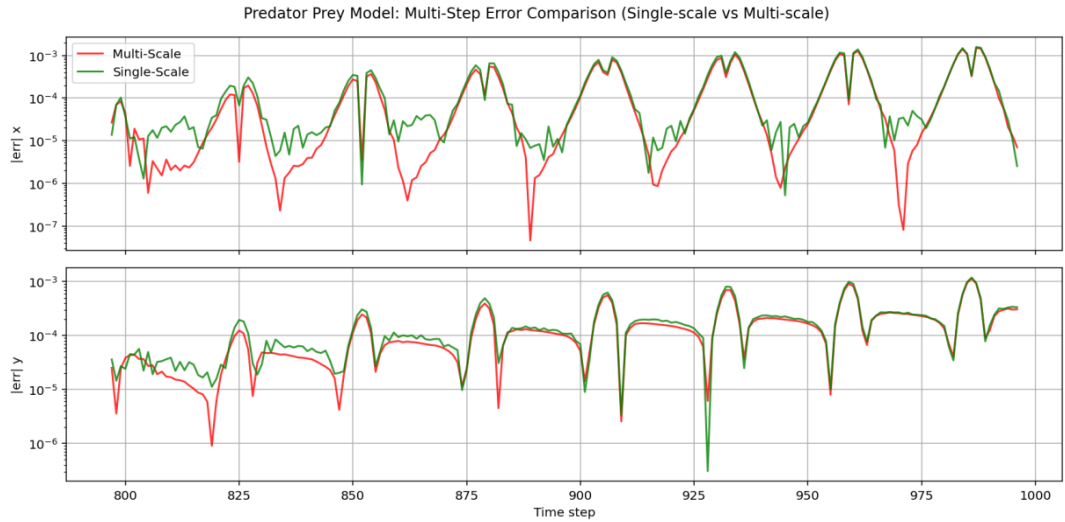

**FIG. 12:** Error Growth in Long-Horizon Predictions: Log-Scale Comparison of Single- and Multi-Scale Random Fourier Feature Reservoirs for predator–prey model

In one-step-ahead prediction (Fig. 10), both Multi-Scale and Single-Scale RFF-RC achieved very small errors. Single-Scale slightly outperformed with $\text{NRMSE}_x = 1.02 \times 10^{-5}$, $\text{NRMSE}_y = 1.00 \times 10^{-5}$, compared to Multi-Scale values of $\text{NRMSE}_x = 1.09 \times 10^{-5}$, $\text{NRMSE}_y = 8.21 \times 10^{-6}$. Thus, for immediate short-term prediction, both approaches were nearly indistinguishable in performance. However, in closed-loop multi-step prediction (Fig. 11), Multi-Scale RFF-RC provided slightly more stable long-horizon tracking. It produced lower errors of $\text{NRMSE}_x = 1.07 \times 10^{-4}$, $\text{NRMSE}_y = 8.23 \times 10^{-5}$, compared to Single-Scale errors of $\text{NRMSE}_x = 1.17 \times 10^{-4}$, $\text{NRMSE}_y = 9.05 \times 10^{-5}$.

The error trajectories (Fig. 12) highlight that while both methods captured the oscillatory prey–predator cycles, Multi-Scale RFF-RC accumulated less error over time. By assigning a narrower bandwidth to the fast prey population ($\sigma_x = 0.1$) and a broader one to the slower predator population ($\sigma_y = 1.0$), Multi-Scale more effectively represented the dual time scales inherent to the system. The Single-Scale RFF-RC, using a single global bandwidth, was unable to balance these two dynamics equally, leading to slightly higher long-term errors.

Overall, the predator–prey results further demonstrate that Multi-Scale RFF-RC retains its advantages beyond neuronal dynamics and can improve predictive stability in ecological systems governed by coupled fast and slow processes.

**E. Morris–Lecar Model**

The Morris–Lecar (ML) model [12] was also used to evaluate the performance of the proposed Multi-Scale Random Fourier Features (RFF-RC) framework. This model is a reduced two-dimensional description of excitable membranes and captures both spiking and bursting neuronal activity. It consists of two coupled differential equations for the membrane potential $V$ and the potassium gating variable $n$. The governing equations are

$$C \frac{dV}{dt} = I_{\text{ext}} - I_{\text{Ca}} - I_K - I_L$$
$$\frac{dn}{dt} = \phi \frac{n_\infty(V) - n}{\tau_n(V)} \qquad (16)$$

where $C$ is the membrane capacitance, $I_{\text{ext}}$ is the applied current, and $I_{\text{Ca}}, I_K, I_L$ are the calcium, potassium, and leak currents. These currents are given by

$$I_{\text{Ca}} = g_{\text{Ca}}\, m_\infty(V)\, (V - V_{\text{Ca}}), I_K = g_K\, n\, (V - V_K), I_L = g_L\, (V - V_L)$$

with maximal conductances $g_{\text{Ca}}, g_K, g_L$ and reversal potentials $V_{\text{Ca}}, V_K, V_L$. The calcium activation is instantaneous and described by

$$m_\infty(V) = \frac{1}{2}\left(1 + \tanh\left(\frac{V - V_1}{V_2}\right)\right)$$

while the potassium activation follows a slower gating process with steady-state value

$$n_\infty(V) = \frac{1}{2}\left(1 + \tanh\left(\frac{V - V_3}{V_4}\right)\right)$$

and voltage-dependent time constant

$$\tau_n(V) = \frac{1}{\cosh\left(\frac{V - V_3}{2V_4}\right)}.$$

The parameter $\phi$ controls the speed of potassium gating relative to voltage. In this setup, $V$ evolves rapidly, while $n$ changes more slowly, producing a natural separation of timescales that is essential for neuronal excitability. By varying the external current $I_{\text{ext}}$ and conductance parameters, the ML model can generate quiescent states, regular spiking, bursting, and even chaotic oscillations. This makes it an excellent test case for evaluating the ability of kernel-based predictors to handle both fast and slow dynamics simultaneously.

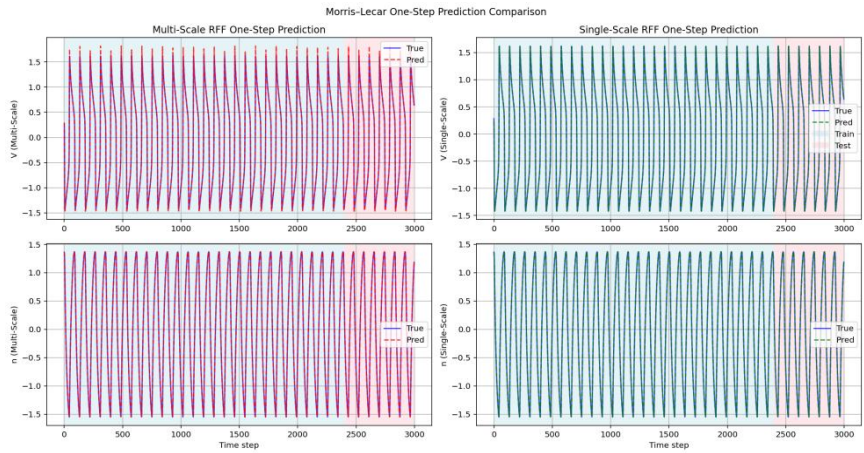

**FIG.13:** One-Step Prediction of Morris–Lecar model Using Reservoir Computing with Single- and Multi-Scale Random Fourier Features

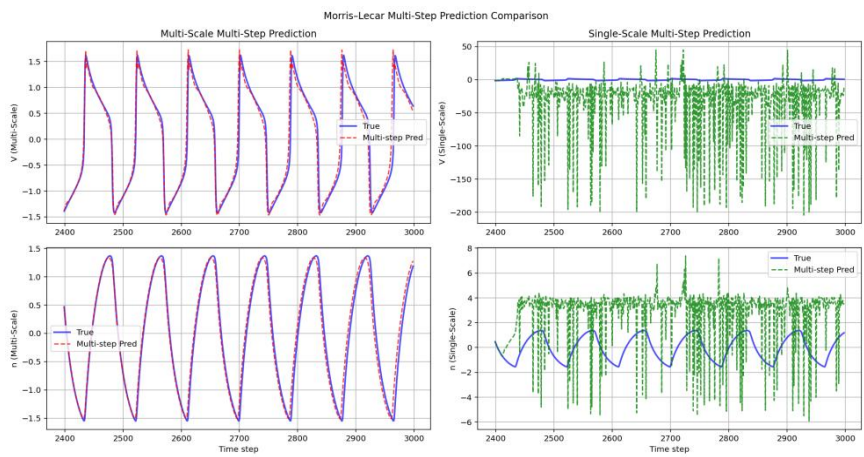

**FIG. 14:** Multi-Step Closed Loop Prediction of Morris–Lecar model Using Reservoir Computing with Single- and Multi-Scale Random Fourier Features

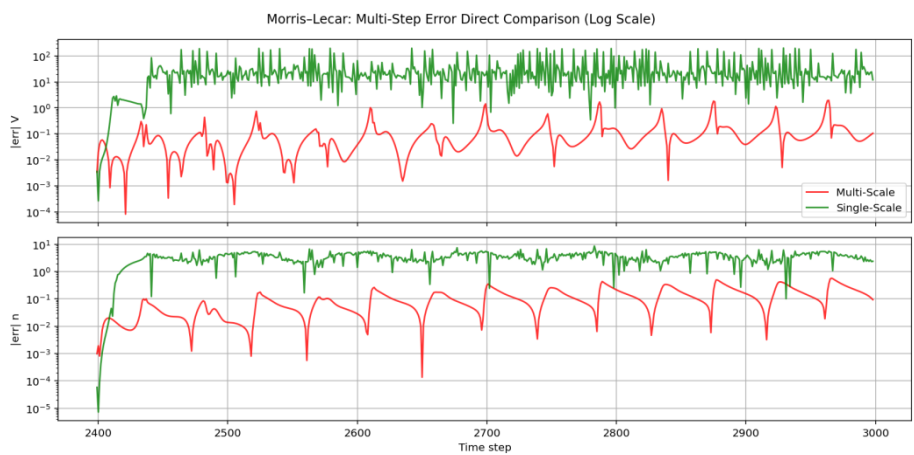

**FIG. 15:** Error Growth in Long-Horizon Predictions: Log-Scale Comparison of Single- and Multi-Scale Random Fourier Feature Reservoirs for Morris–Lecar model

In one-step-ahead prediction (Fig. 13), the Single-Scale RFF-RC provided slightly lower errors, with $\text{NRMSE}_V = 8.91 \times 10^{-3}$ and $\text{NRMSE}_n = 2.77 \times 10^{-4}$. By contrast, the Multi-Scale RFF-RC, using variable-specific bandwidths ($\sigma_V = 1, \sigma_n = 10$), achieved $\text{NRMSE}_V = 1.89 \times 10^{-2}$ and $\text{NRMSE}_n = 7.68 \times 10^{-4}$. However, in multi-step closed-loop prediction (Fig. 14), the superiority of Multi-Scale RFF-RC became striking. The Single-Scale framework showed rapid error growth and eventually diverged ($\text{NRMSE}_V = 1.90 \times 10^1$, $\text{NRMSE}_n = 1.27$). In contrast, Multi-Scale RFF-RC maintained stability and accurate tracking with much smaller errors ($\text{NRMSE}_V = 8.91 \times 10^{-2}$, $\text{NRMSE}_n = 5.48 \times 10^{-2}$).

Error trajectories (Fig. 15) confirmed these observations. Multi-Scale RFF-RC slowed divergence, reduced fluctuations, and maintained reliable tracking of both the fast membrane potential and the slower potassium gating variable across long horizons. The advantage comes from its ability to apply distinct bandwidths for different variables: a narrower kernel for the fast voltage oscillations and a broader kernel for the slower gating dynamics. A single global bandwidth, by contrast, cannot balance these requirements and tends to overfit one timescale while underrepresenting the other. As a result, although Single-Scale RFF-RC achieved slightly better accuracy in one-step prediction, Multi-Scale RFF-RC offered far superior long-term stability and robustness, which is crucial for excitable systems like the Morris–Lecar model.

### F. Ricker map

The Ricker map [13] with seasonal forcing was also studied to test the generality of the Multi-Scale Random Fourier Features (RFF) framework beyond neuronal dynamics. The Ricker model originates from population ecology and describes density-dependent growth with saturation effects. It can be extended to a fast–slow formulation by coupling the population dynamics with a slowly varying growth rate. The system considered here is defined by

$$\begin{aligned} x_{n+1} &= x_n \exp\left(r_{n+1}\left(1 - \frac{x_n}{K}\right)\right) \\ r_{n+1} &= r_n + \epsilon\left(\mu - r_n + \alpha \sin\left(\frac{2\pi n}{T}\right)\right) \end{aligned} \quad (17)$$

where $x_n$ represents the population at generation $n$, $r_n$ is the slowly varying intrinsic growth rate, $K$ is the carrying capacity, $\mu$ is the baseline growth rate, and $\alpha$ and $T$ control seasonal forcing. The parameter $\epsilon \ll 1$ sets the separation of timescales, making $x_n$ the fast ecological variable and $r_n$ the slow forcing component. This model captures ecological scenarios where populations evolve under density dependence while growth rates drift due to environmental changes or seasonal effects.

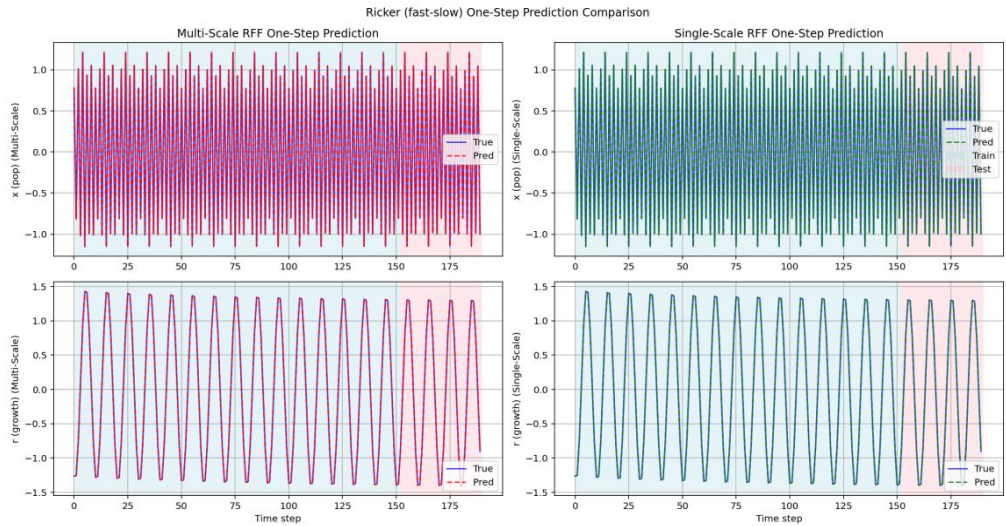

**FIG.16:** One-Step Prediction of Morris–Lecar model Using Reservoir Computing with Single- and Multi-Scale Random Fourier Features

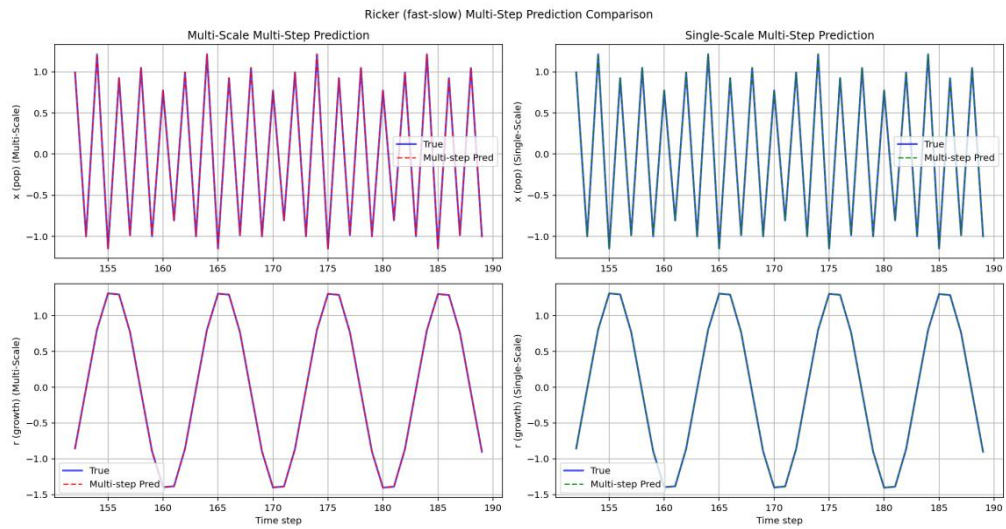

**FIG. 17:** Multi-Step Closed Loop Prediction of Morris–Lecar model Using Reservoir Computing with Single- and Multi-Scale Random Fourier Features

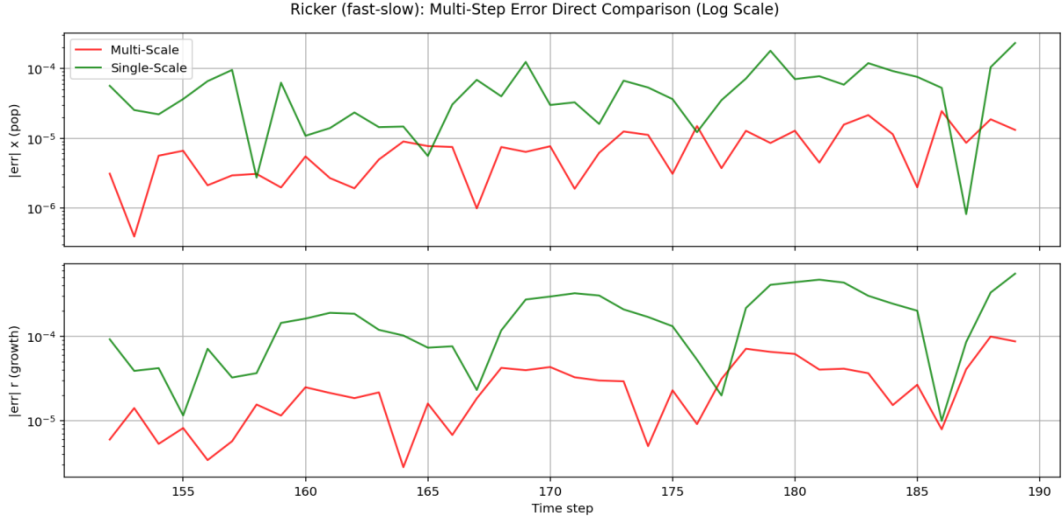

**FIG. 18:** Error Growth in Long-Horizon Predictions: Log-Scale Comparison of Single- and Multi-Scale Random Fourier Feature Reservoirs for Morris–Lecar model

In one-step-ahead prediction (Fig. 16), Multi-Scale RFF-RC clearly outperformed Single-Scale. Using variable-specific bandwidths ($\sigma_x = 1.0$, $\sigma_r = 10.0$), it achieved $\text{NRMSE}_x = 2.55 \times 10^{-6}$ and $\text{NRMSE}_r = 6.13 \times 10^{-6}$. Single-Scale, with a global bandwidth, produced significantly higher errors of $\text{NRMSE}_x = 2.52 \times 10^{-5}$ and $\text{NRMSE}_r = 4.62 \times 10^{-5}$. Thus, the multi-scale setup improved accuracy by nearly an order of magnitude.

For multi-step closed-loop prediction (Fig. 17), Multi-Scale again maintained superior performance. It achieved $\text{NRMSE}_x = 4.10 \times 10^{-6}$ and $\text{NRMSE}_r = 1.35 \times 10^{-5}$, compared to Single-Scale values of $\text{NRMSE}_x = 3.13 \times 10^{-5}$ and $\text{NRMSE}_r = 8.63 \times 10^{-5}$. The differences in error accumulation were substantial, with Single-Scale diverging more quickly, particularly in the slow forcing variable.

Error trajectories (Fig. 18) confirmed that Multi-Scale RFF-RC preserved both the short-term fluctuations of population dynamics and the longer-term seasonal drift of the growth rate. By assigning narrower bandwidths to the population variable and broader bandwidths to the slowly varying growth rate, the framework was able to capture ecological variability across different time scales. In contrast, a single global bandwidth failed to represent both fast and slow dynamics simultaneously.

**IV. Conclusion**

In this study, we developed a reservoir computing framework that integrates delay embedding with random Fourier feature (RFF-RC) mappings and systematically compared two formulations: a single-scale RFF-RC reservoir, which uses a fixed kernel bandwidth, and a multi-scale RFF-RC reservoir, which combines multiple bandwidths to capture both fast and slow temporal dependencies. Applying the framework to six benchmark systems—the Rulkov map, Morris–Lecar model, Hindmarsh–Rose model, Izhikevich model, Ricker map, and Predator–Prey dynamics—we found that forecasting performance depends critically on the ability to represent multi-scale structures. While the single-scale RFF-RC reservoir achieves competitive short-term accuracy, it fails to sustain predictive stability in long-horizon closed-loop forecasts. By contrast, the multi-scale RFF-RC reservoir creates a richer implicit feature map, simultaneously encoding rapid transients and slow modulatory processes, thereby yielding consistently lower error and improved robustness.

The consistent gains across neuronal and ecological models demonstrate that the multi-scale formulation is not only numerically more accurate but also structurally better aligned with the intrinsic fast–slow interactions that drive system behavior. More broadly, this approach highlights the value of explicitly incorporating multi-scale feature mappings into reservoir computing for modeling nonlinear dynamics with timescale separation. Looking ahead, future work will extend the framework to real-world data, adaptive prediction, and advanced reservoir designs, paving the way for scalable and reliable forecasting of multi-scale phenomena across biology, ecology, physics, and engineering.

## References

1. Kuehn, C. (2015). *Multiple time scale dynamics* (Vol. 191). Springer.
2. Murray, J. D. (2007). *Mathematical biology: I. An introduction* (Vol. 17). Springer.
3. Vlachos, D. G. (2005). A review of multiscale analysis: Examples from systems biology, materials engineering, and other fluid–surface interacting systems. *Advances in Chemical Engineering, 30*, 1–61.
4. Lei, J. (2013). Slow–fast dynamics. In *Encyclopedia of Systems Biology* (pp. 1955–1956). Springer.
5. Walpole, J., Papin, J. A., & Peirce, S. M. (2013). Multiscale computational models of complex biological systems. *Annual Review of Biomedical Engineering, 15*(1), 137–154.
6. Meier-Schellersheim, M., Fraser, I. D. C., & Klauschen, F. (2009). Multiscale modeling for biologists. *Wiley Interdisciplinary Reviews: Systems Biology and Medicine, 1*(1), 4–14.
7. Weinan, E. (2011). *Principles of multiscale modeling*. Cambridge University Press.
8. Scheffer, M., & Carpenter, S. R. (2003). Catastrophic regime shifts in ecosystems: Linking theory to observation. *Trends in Ecology & Evolution, 18*(12), 648–656.
9. Rulkov, N. F. (2002). Modeling of spiking–bursting neural behavior using two-dimensional map. *Physical Review E, 65*(4), 041922.
10. Izhikevich, E. M. (2003). Simple model of spiking neurons. *IEEE Transactions on Neural Networks, 14*(6), 1569–1572.
11. Hindmarsh, J. L., & Rose, R. M. (1984). A model of neuronal bursting using three coupled first order differential equations. *Proceedings of the Royal Society of London. Series B. Biological Sciences, 221*(1222), 87–102.

12. Morris, C., & Lecar, H. (1981). Voltage oscillations in the barnacle giant muscle fiber. *Biophysical Journal, 35*(1), 193–213.

13. Ricker, W. E. (1954). Stock and recruitment. *Journal of the Fisheries Board of Canada, 11*(5), 559–623.

14. Jaeger, H. (2001). The "echo state" approach to analysing and training recurrent neural networks—with an erratum note. *GMD Technical Report, 148*, 34.

15. Lukoševičius, M., & Jaeger, H. (2009). Reservoir computing approaches to recurrent neural network training. *Computer Science Review, 3*(3), 127–149.

16. Lukoševičius, M. (2012). A practical guide to applying echo state networks. In G. Montavon, G. B. Orr, & K.-R. Müller (Eds.), *Neural Networks: Tricks of the Trade* (pp. 659–686). Springer.

17. Gauthier, D. J. (2018). Reservoir computing: Harnessing a universal dynamical system. *SIAM News, 51*(12).

18. Bollt, E. (2021). On explaining the surprising success of reservoir computing forecaster of chaos? The universal machine learning dynamical system with contrast to VAR and DMD. *Chaos: An Interdisciplinary Journal of Nonlinear Science, 31*(1), 013108.

19. Gauthier, D. J., Bollt, E., Griffith, A., & Barbosa, W. A. S. (2021). Next generation reservoir computing. *Nature Communications, 12*(1), 5564.

20. Pathak, J., Hunt, B., Girvan, M., Lu, Z., & Ott, E. (2018). Model-free prediction of large spatiotemporally chaotic systems from data: A reservoir computing approach. *Physical Review Letters, 120*(2), 024102.

21. Pathak, J., Wikner, A., Fussell, R., Chandra, S., Hunt, B. R., Girvan, M., & Ott, E. (2018). Hybrid forecasting of chaotic processes: Using machine learning in conjunction with a knowledge-based model. *Chaos: An Interdisciplinary Journal of Nonlinear Science, 28*(4), 041101.

22. Lu, Z., Hunt, B. R., & Ott, E. (2018). Attractor reconstruction by machine learning. *Chaos: An Interdisciplinary Journal of Nonlinear Science, 28*(6), 061104.

23. Chattopadhyay, A., Hassanzadeh, P., & Subramanian, D. (2020). Data-driven prediction of a multi-scale Lorenz 96 chaotic system using deep learning methods: Reservoir computing, ANN, and RNN-LSTM. *Nonlinear Processes in Geophysics Discussions*, 1–26.

24. Pandey, S., & Schumacher, J. (2020). Reservoir computing model of two-dimensional turbulent convection. *Physical Review Fluids, 5*(11), 113506.

25. Nakai, K., & Saiki, Y. (2018). Machine-learning inference of fluid variables from data using reservoir computing. *Physical Review E, 98*(2), 023111.

26. Kobayashi, M. U., Nakai, K., Saiki, Y., & Tsutsumi, N. (2021). Dynamical system analysis of a data-driven model constructed by reservoir computing. *Physical Review E, 104*(4), 044215.

27. Gallicchio, C., & Micheli, A. (2017). Echo state property of deep reservoir computing networks. *Cognitive Computation, 9*(3), 337–350.

28. Gallicchio, C., Micheli, A., & Pedrelli, L. (2018). Deep reservoir computing: A critical experimental analysis. *Neurocomputing, 268*, 87–99.

29. Gao, R., et al. (2021). Time series forecasting based on echo state network and empirical wavelet transformation. *Applied Soft Computing, 102*, 107111.

30. Zheng, K., et al. (2020). Long-short term echo state network for time series prediction. *IEEE Access, 8*, 91961–91974.

31. Tanaka, G., et al. (2022). Reservoir computing with diverse timescales for prediction of multiscale dynamics. *Physical Review Research, 4*(3), L032014.

32. Parlitz, U. (2024). Learning from the past: Reservoir computing using delayed variables. *Frontiers in Applied Mathematics and Statistics, 10*, 1221051.

33. Kong, L.-W., Brewer, G. A., & Lai, Y.-C. (2024). Reservoir-computing based associative memory and itinerancy for complex dynamical attractors. *Nature Communications, 15*(1), 4840.

34. Ebato, Y., et al. (2024). Impact of time-history terms on reservoir dynamics and prediction accuracy in echo state networks. *Scientific Reports, 14*(1), 8631.

35. Lin, Z., et al. (2024). Learning noise-induced transitions by multi-scaling reservoir computing. *Nature Communications, 15*(1), 6584.

36. Tsuchiyama, K., et al. (2023). Effect of temporal resolution on the reproduction of chaotic dynamics via reservoir computing. *Chaos: An Interdisciplinary Journal of Nonlinear Science, 33*(6), 063118.

37. Lin, Z., Kurths, J., & Tang, Y. (2025). Multi-scaling reservoir computing learns noise-induced transitions with Lévy noise. *Chaos: An Interdisciplinary Journal of Nonlinear Science, 35*(7), 073124.

38. Grigoryeva, L., Lim Jing Ting, H., & Ortega, J.-P. (2025). Infinite-dimensional next-generation reservoir computing. *Physical Review E, 111*(3), 035305.

39. Danilenko, G. O., et al. (2025). Reservoir computing with state-dependent time delay. *Physical Review E, 111*(3), 035313.

40. Du, Y., et al. (2025). Multifunctional reservoir computing. *Physical Review E, 111*(3), 035303.

41. Inoue, S., et al. (2024). Multi-scale dynamics by adjusting the leaking rate to enhance the performance of deep echo state networks. *Frontiers in Artificial Intelligence, 7*, 1397915.3.

42. Rahimi, A., & Recht, B. (2007). Random features for large-scale kernel machines. In *Advances in Neural Information Processing Systems* (Vol. 20).